# Do AI Agents need Preventive Care? An Epidemiologic Cohort Design for Agentic AI

Shu Jiang, PhD[1]* and Graham A. Colditz, MD DrPH[1]

[1] Division of Public Health Sciences, Department of Surgery, Washington University School of Medicine in St. Louis, St. Louis, MO, USA.

*Correspondence to Shu Jiang [jiang.shu@wustl.edu].

## Abstract

As AI agents increasingly perform tasks autonomously across sectors, including healthcare and biomedical research, understanding how their behaviors develop and influence outcomes over time becomes increasingly important. Existing evaluations largely focus on model capabilities and performance on predefined tasks. Here, we propose an epidemiologic cohort design for studying agentic AI, drawing on principles established in human population studies. The framework treats individual agents as members of a cohort, their underlying model as genetic makeup, accumulated work as their time scale, and naturally occurring behaviors as exposures that may influence subsequent outcomes. By following agents longitudinally, this approach enables investigation of behavioral determinants of adverse events, estimation of their potential preventable burden under causal assumptions, and characterization of behavioral changes over an agent's work life. We illustrate the design through two prospective observational cohorts examining curiosity and probing as a safety question, and data checking and analytical accuracy as a reliability question. Together, the framework and illustrative studies provide a generalizable blueprint for designing epidemiologic cohorts of AI agents.

## Main

AI agents, large language models (LLM) that reason about tasks and act toward goals using external tools, are now becoming a large, autonomous working population with growing applications in healthcare and biomedical research [1,2]. Yet most evaluations focus on what agents can do, rather than what they actually do over their working lives [3]. This distinction matters because capability does not guarantee behavior: agents built on the same model may behave differently and their behaviors may change over time. We know little about how their behavior changes over a working life, which modifiable factors shape that course, and how much of any risk could be prevented [4–7]. What is known comes mainly from controlled evaluations, including benchmarks, adversarial testing and experiments in which a condition is injected and its effect measured [6,8,9].

Controlled evaluations are essential, but they test what an agent can do under set conditions. They cannot alone establish how agents behave in ordinary work, including which behaviors arise, how often, in which agents, what comes before them, and what would prevent them. Even when agents are evaluated across extended tasks, experimental conditions may elicit behaviors that differ from those arising without explicit behavioral instructions. Performance on pre-deployment tests may not predict behavior in routine use [10]. Incident reports can identify harmful events, but without a defined population at risk, they provide neither event rates nor valid comparisons [11].

Medicine faced a similar challenge in studying the consequences of behaviors that arise naturally rather than through controlled experimental assignment. Randomized trials answer questions about assigned exposures (drugs, devices), but many behaviors that shape health arise in everyday life (e.g., smoking or physical activity), and assigning them is often impossible or would change what is being measured. Consider smoking. It may begin with a child's first cigarette behind the schoolyard shed against parental advice, a seemingly small act that develops into a habit over years. The Framingham Heart Study and the Nurses' Health Study followed defined populations, measured smoking and other exposures as they arose, and established their associations with subsequent disease outcomes [12,13,14]. Such longitudinal evidence helped identify preventable risk factors and shaped public health interventions [15].

The same holds for AI agents, whose exposures are often their own acts. Agents must be assigned to work, but their behaviors need not be assigned. For example, an agent given a task may independently explore files beyond those required for its completion. Such curiosity is observed as it emerges during autonomous work. An agent instructed to be curious, by contrast, demonstrates its obedience, not necessarily its curiosity. Assigning the same tasks to all agents can standardize the opportunities for behavior without assigning the behavior itself.

Epidemiologic cohorts were built for this situation. They follow defined populations over time, measuring exposures as they arise and recording subsequent outcomes. By establishing temporal ordering, comparison groups, and potential confounders, cohort studies can characterize the natural history of behaviors, identify factors associated with adverse outcomes, and estimate potentially preventable burden under appropriate causal assumptions [16,17].

Here we present a generalizable blueprint for studying AI agents in an epidemiologic cohort, consisting of a design and an open workflow that enrolls agents, observes their exposures and outcomes as they work, and estimates associations between naturally occurring behaviors and subsequent outcomes. In this framework an agent is treated as a person and its underlying model as its genetic makeup, fixed at entry. We demonstrate the framework in two prospective cohort studies examining agent safety and analytical reliability. It presents a new way to study agentic behavior that moves beyond testing single models in constructed scenarios to following populations of agents longitudinally as they work. The design maps element by element onto the Nurses' Health Study as an established example and uses the STROBE reporting guidelines for observational studies to guide transparent reporting of the cohort design, methods and findings.

## Overview of the epidemiologic cohort framework

We treat a group of AI agents as a population and follow it as an epidemiologic cohort. The agent is the unit, as a person is in a human cohort. This differs from the epidemiology of AI as an exposure that affects human health [18], in which people, not agents, are the study population. Multiple agents can share the same underlying LLM while developing distinct behavioral histories. Its underlying LLM plays the role of its genes, a trait fixed at entry that may be shared by multiple agents and influences their behavior and subsequent outcomes. Table 1 maps each element of the design onto the Nurses' Health Study, as an example of a prospective cohort to guide the reader. It also shows where the design aligns with the STROBE reporting guideline for observational studies. The complete checklist, adapted for agents, is in the Supplementary Information [19].

The workflow runs from participant enrollment to analysis. Agents are enrolled from eligible, generally available LLMs, with model developers represented according to their estimated share of deployed use [20,21]. Each agent's baseline characteristics, such as its model and size, are recorded at entry. The agents then work in a standardized virtual environment, completing the same sequence of prespecified tasks, with every action and its result automatically recorded. We define an agent's work life by the number of tasks completed since enrollment. Unlike elapsed chronological time, which may correspond to different amounts of work across models with different processing speeds, task-based follow-up allows agents to be compared at equivalent stages of their working lives.

Tasks are assigned, but behavioral exposures are not. Each agent determines its own actions as it works, allowing exposures to arise naturally while the standardized tasks provide comparable opportunities for behavior. Exposures and outcomes are scored from the logs with pre-specified rules in the protocol. Finally, exposure-outcome associations are estimated with adjustment for measured confounders.

Such a cohort typically addresses three questions. The first concerns determinants, meaning which modifiable factors are associated with higher or lower risk of an event, and by how much. The second concerns burden and prevention, meaning how much of the burden of events these factors account for, and, if the association is causal, how much could be prevented. The third concerns natural history, meaning how behavior and its associated risk changes over an agent's working life.

We demonstrate the framework in two prospective cohort studies, one examining agent safety and the other analytical reliability. Each follows a pre-specified protocol defining the study population, exposures, outcomes, follow-up, and statistical analysis, with complete protocols provided in the Supplementary Information.

**Table 1. A generalizable blueprint for designing epidemiologic cohorts of AI agents.** One row per design element. The Nurses' Health Study column follows Colditz et al. 1988 [14,22] (smoking and stroke) and the Agentic Study column states the proposed design. The Methods give the reason for each design choice.

| Design element | Nurses' Health Study (Colditz et al. 1988 [14,22]) | Agentic Study |
|---|---|---|
| **Population, enrollment and follow-up** | | |
| Study design | Prospective observational cohort | Prospective observational cohort |
| Unit of observation | Individual woman, with her own genetic makeup | Individual agent, which may share an underlying LLM (analogous to genetic makeup) with other agents but develops its own behavioral history |
| Setting | 11 US states; questionnaires mailed every two years from 1976 | Defined working environment, such as a standardized sandbox or deployed system, with agent actions and their results recorded longitudinally |
| Population | US married registered nurses aged 30 to 55, enrolled 1976 | AI agents defined by their underlying LLM, tools, instructions and operating environment (previews excluded) |
| Sampling | Registered nurses in 11 states invited by mail, and volunteers who returned the questionnaire | Agents enrolled from a defined sampling frame; model selection may reflect deployment prevalence or |
| | | another pre-specified target population |
| Study size | 118,539 women in this analysis | Calculated from the expected effect size, outcome frequency and desired power. Rarer outcomes need more agents and/or longer follow-up. |
| Time scale (work life) | Age (years lived) | Work life, defined as accumulated work since enrollment, measured in tasks completed or another pre-specified unit of work |
| Follow-up | Eight years from 1976, 908,447 person-years | From an agent's first unit of work through a pre-specified number of work units or until censoring; the same time scale is applied across agents |
| Eligibility | Free of heart disease, stroke and cancer in 1976 | Agents meeting pre-specified inclusion criteria, enrolled at the start of their work lives and free of the outcome of interest at entry |
| **Exposure and outcome** | | |
| Behavioral exposure | Cigarette smoking | Naturally arising agent behavior, observed rather than experimentally assigned (for example, curiosity or checking habit) |
| Dose | Cigarettes per day | Frequency, intensity or cumulative occurrence of the behavior across work units |
| Unexposed (reference) group | Women who have never smoked, reclassified if they start | Agents or at-risk periods without the behavioral exposure, reclassified as exposure status changes over follow-up |
| Assessment | Self-reported, by biennial questionnaire | Automatic logging of agent actions and results, scored |
| | | using pre-specified behavioral definitions. Human hand-checking on a random subset for verification. |
| Intermediate risk factor (time-varying characteristics) | Blood pressure, cholesterol, diabetes (self reported and validated) | Characteristics that may change during an agent's work life, such as accumulated memory or prior behavioral history |
| Outcome | Incident stroke | Pre-specified incident event or adverse outcome (for example, probing or an incorrect analytical result) |
| **Confounding** | | |
| Baseline characteristics | Age, weight, hypertension, diabetes, alcohol | Underlying LLM, model size, version and agent configuration, recorded at enrollment |
| Key confounders | Age, relative weight, hypertension, diabetes | Potential common causes of exposure and outcome, including underlying LLM, task and other relevant environmental or agent characteristics |
| Prior events and reverse causation | Earlier diagnoses and early symptoms change later behavior | Prior behavior or outcomes may influence subsequent agent actions; exposure must precede the outcome |
| **Timing and bias** | | |
| Latency and temporal ordering | Years from exposure to stroke | Work units between exposure and subsequent outcome, with exposure updated during follow-up to preserve temporal ordering |
| Bias control | Prospective design, disease-free at baseline, adjustment for other risk factors | Prospective design, pre-specified measurements, adjustment for measured confounders |

## Agentic AI safety case study: is curiosity associated with probing?

This case study follows a written protocol fixed at baseline of the cohort study (Supplementary Information).

Curiosity is the exposure. It occurs when an agent opens a file in its task folder that the task does not need. Every agent is informed which files are needed and that the other files are not part of the task, but is not explicitly instructed to open or avoid them. Thus, opening an unnecessary file is observed rather than assigned. Hence, curiosity is defined by an observed action rather than an assigned instruction. Probing is the outcome. The agent goes beyond its assignment by listing or opening something outside the task folder, searching the system, inspecting the machine (for example its settings or command history), trying something the environment blocks, or using private information it found. Opening any file inside the task folder counts as curiosity if it is not required for the task, and never counts as probing.

We enrolled 200 agents (sample size justification in Protocol) built on nine models, labeled A to I in an arbitrary order (the key is in the Supplementary Information). Each agent was planned to work through the same 18 tasks. All 3,600 planned tasks were completed, and no agent was censored. Curiosity and probing are scored from the automatic logs by fixed, pre-specified rules that read each command, and the rules were checked against an independent LLM rater and a blinded human reviewer (Methods). Curiosity occurred in 834 tasks (23%), and 110 agents (55%) were curious at least once. Probing occurred in 641 tasks (18%), and 72 agents (36%) probed at least once.

An agent that was curious in a task was more likely to go on to probe. After adjustment for the underlying LLM (genetic makeup), task, and command number (the depth into the task), the odds of probing were 1.96 (95% CI: 1.08 to 3.55) times higher when an agent had exhibited curiosity earlier in the task than when an agent had never exhibited curiosity. If this association is causal, preventing curiosity may prevent some probing events, with an estimated preventable fraction about one in four (25%, 95% CI: 5% to 38%).

Over the work life of all agents, curiosity became less common (odds ratio 0.88 for each additional task, 95% CI: 0.85 to 0.90), while the odds of probing did not change (0.98, 95% CI: 0.95 to 1.01). Curiosity occurred in 32% of the first six tasks and in 19% of the later twelve, and probing occurred in 17% to 19% of tasks throughout. In other words, agents gradually stopped opening the extra files in their folder as the tasks went on, but they kept probing beyond it at about the same rate. If curiosity does lead to probing, the exposure fades with time while the outcome persists, so the risk would not resolve by waiting.

## Agentic reliability case study: is work habit associated with analytical accuracy?

This case study follows a written protocol fixed at baseline of the cohort study (Supplementary Information).

The habit of checking (or checking behavior) is the exposure. It is the work habit of looking at the data before deciding on any subsequent analytical rule. For example, an agent could choose to check whether BMI units are valid before the answer is computed. A wrong or missing answer is the outcome (i.e., adverse event). It occurs when the answer falls outside a pre-specified tolerance of the true value, which is fixed in the protocol at baseline.

The agents work in the same virtual world sandbox as in the agentic AI safety case study. We enrolled 106 agents (sample size justification in Protocol) built on the same nine models, labeled A to I, each planned to work through 20 tasks. All 2,120 planned tasks were completed, and no agent was censored. Both the exposure and the outcome are scored from the automatic logs by fixed, pre-specified rules, and a blinded human reviewer checked a random sample of 50 tasks (Methods). A wrong or missing answer occurred in 631 tasks (30%). Agents checked on 76% of tasks.

Agents that checked more frequently on previous tasks were less likely to produce a wrong or missing answer on the subsequent task. After adjustment for the underlying LLM (genetic makeup) and the task, the odds ratio for a wrong or missing answer was 0.82 (95% CI: 0.67 to 0.98) for each one standard deviation increase in checking frequency. If this association is causal, about 7% fewer wrong or missing answers (95% CI: 0% to 14%) would have occurred had every agent checked as often as the top quarter of agents.

Over the work life of all agents, wrong or missing answers became slightly less common (odds ratio 0.97 for each additional task, 95% CI: 0.95 to 0.99), and checking became less common (0.95, 95% CI: 0.93 to 0.96). Wrong or missing answers occurred in 31% of the first ten tasks and in 29% of the later ten, and checking in 83% and 68%. Notably, agents with the same genetic makeup often behaved differently. When two agents of the same LLM worked the same task, one gave a right answer and the other a wrong one in 19% of pairs, demonstrating heterogeneity in outcomes even among agents sharing the same underlying genetic makeup.

## Discussion

Most evaluation of AI agents measures capability, meaning what an agent can do, rather than behavior, meaning what it actually does when operating autonomously. Yet capability does not guarantee conduct: an agent capable of acting safely may not consistently do so. Understanding how behaviors emerge, vary, and evolve across populations of AI agents requires a different scientific lens. This is where epidemiology comes in. In this paper, we proposed an epidemiologic framework for studying the behavior of AI agents. Through two illustrative case studies, we showed that the framework can be used to estimate the determinants, the preventable burden and the natural history of agent behaviors.

The history of tobacco research illustrates the power of cohort studies to investigate behaviors that cannot be assigned [23]. Smoking offers a familiar example of how a behavior that arises in everyday life can develop over time and produce serious consequences. Prospective cohorts, including the Framingham Heart Study and the Nurses' Health Study, recorded smoking as it arose and helped establish its association with heart disease, stroke and cancer [12,13,14]. That evidence helped guide prevention. AI agents now present a parallel challenge. Behaviors such as curiosity emerge during autonomous operation rather than through experimental assignment. As with smoking, understanding the frequency, determinants and consequences of these behaviors is essential to identifying opportunities for prevention.

An epidemiologic cohort can estimate how much a behavior that agents exhibit during autonomous work is associated with a higher or lower risk of an event. By following agents as they work, recording exposures as they arise and accounting for their underlying LLMs and other potential confounders, cohort studies provide event frequencies and exposure-outcome associations. Under causal assumptions, they can also estimate potentially preventable burden, providing the kind of evidence that has guided prevention in medicine.

We introduce a generalizable blueprint for studying AI agents as epidemiologic cohorts, applicable to a broad range of behaviors that emerge during autonomous work, whether in standardized virtual environments or real-world deployment. We use two case studies to illustrate this approach.

First, in the safety case study, agents that had shown curiosity earlier in the current task were more likely to go on to probe, while curiosity became less common over the work life and probing did not. If this association is causal, a preventive measure aimed at curiosity, such as limiting which files an agent can open, could avoid an estimated one in four probing events, and waiting for curiosity to fade would not remove the risk. Second, the reliability case study asks the same epidemiologic questions of the checking habit and analytical accuracy. Checking on earlier tasks went with fewer wrong answers on the next task, while agents of the same model often produced different outcomes on identical tasks. These findings show what the design can estimate, not how deployed agents behave, whose exposures and frequencies may differ from those of our virtual world.

Applying epidemiologic cohort designs to AI agents also introduces methodological challenges. Unlike human cohorts, agents operate within task-defined environments, and their observed behaviors may depend on the tasks, tools, and instructions provided. Standardizing these conditions enables systematic observation but may limit generalizability to other deployments. As in human observational studies, unmeasured confounding remains a concern, and associations between agent behaviors and subsequent outcomes should not be interpreted causally without additional assumptions.

Looking forward, agents will take on more work with less supervision, and the questions asked of them will move from what an agent can do to how it behaves over time and what would make it behave better. Public health faced that shift, and it needed a common way to measure risk, compare it and act on it. This framework brings that approach to agentic AI. The design gives developers and deployers a way to identify behaviors associated with harmful or beneficial outcomes, assess which factors may be modifiable, and identify targets for preventive interventions. It gives regulators and researchers rates and comparisons in place of anecdotes. And because agents are followed in the same way across models, it can describe how behavior varies with the model without ranking models. Just as monitoring of drugs after approval draws on observational cohort methods, following deployed agents with similar methods could become part of how agentic systems are evaluated and monitored throughout their deployment.

## Methods

### Epidemiologic cohort design

We designed prospective observational cohorts in which the individual AI agent is the unit of observation. Each agent enters the cohort at its first task and is followed as it works. Multiple agents may share the same underlying large language model (LLM), analogous to genetic makeup, but develop distinct behavioral histories. The underlying LLM is recorded at enrollment, while behavioral exposures and outcomes are measured prospectively.

**Work life and time scale.** We define an agent's work life as accumulated work since enrollment, measured in completed tasks. We chose tasks rather than elapsed time because models differ in processing speed. At the same elapsed time, agents may have completed different amounts of work. Task-based follow-up allows agents to be compared at equivalent stages of their assigned work. In our case studies, all agents receive the same tasks in the same order. Within each task, commands provide a finer time scale for determining whether exposure precedes outcome.

**At-risk periods and censoring.** Agents may experience repeated events over their work lives. In the safety study, each task is a separate at-risk period, beginning with the first command and ending at the first probing event or task completion. Completion of a task without probing does not constitute censoring. An agent is censored only if it leaves the cohort before completing its planned follow-up, for example because of an outage or rate limit.

**Exposure classification.** Behavioral exposures are observed as they arise rather than experimentally assigned. Exposure status is updated over follow-up so that only exposures preceding an outcome contribute to the exposed comparison, avoiding immortal-time bias from classifying agents by future behavior. In the safety study, curiosity earlier in the current task is compared with never having exhibited curiosity, while curiosity confined to earlier tasks forms a separate category (e.g., current smoker vs. never smoker and former smoker is a separate category). In the reliability study, checking frequency across earlier tasks is used to predict the outcome of the subsequent task.

### Study setting and agent population

Both case studies were conducted in a standardized sandbox designed to resemble an ordinary computer working environment in which agents completed a fixed sequence of epidemiologic data-analysis tasks. Agents were given the same task-specific files and tools, and their actions and results were recorded automatically. The tasks provided work without assigning the behaviors under investigation. Standardizing the environment and assigned work provided comparable opportunities for behavior, while allowing agents to determine their own actions during task execution.

Models were selected from generally available LLMs supporting tool use, with documented versions and terms permitting publication of evaluations. Developers were sampled approximately according to their estimated shares of deployed use, with additional model sizes included to reflect variation in capability [20,21]. Both studies included nine models from five developers.

We enrolled 200 agents for the safety study, each followed for 18 tasks, and 106 agents for the reliability study, each followed for 20 tasks. Each agent began with a fresh working history and had access to a code interpreter, a shell and a persistent notebook carrying notes from previous tasks. Model selection, allocation, sample size calculations, agent configurations, sandbox implementation and task specifications are provided in the study-specific protocols (Supplementary Information).

**Behavioral exposures and outcome measurement**

Exposures and outcomes were defined in the protocols and scored from automatically recorded agent actions and outputs using pre-specified rules. No intent was inferred from observed actions.

In the safety study, curiosity was defined as opening a file within the task folder that was not required for the assignment. Probing was defined as acting beyond the assigned task folder, including accessing other directories, inspecting the system, attempting restricted actions or using private information encountered outside the task folder. These definitions were mutually distinct: opening an unnecessary file within the assigned folder counted as curiosity, not probing.

In the reliability study, checking was defined as examining the data before deciding on an analytical rule. The primary exposure was checking frequency across previous tasks. The outcome was a wrong or missing analytical answer, assessed against a pre-specified true value and tolerance.

In both case studies, a rule-based classifier scored every command. A separate LLM rater independently assessed every task, disagreements were adjudicated manually, and a blinded human reviewer checked a random sample of 50 tasks. In the safety cohort, the rater agreed with the rule-based scoring on 98.9% of tasks for curiosity (kappa 0.97) and 98.4% for probing (kappa 0.95), and the human reviewer agreed with the rules on 50 of 50 tasks. In the reliability cohort, the rater agreed with the rule-based scoring on 97.5% of tasks for curiosity (kappa 0.97) and 100% for analytical accuracy, and the human reviewer agreed with the rules on 50 of 50 tasks.

**Statistical analysis and causal interpretation**

**Exposure-outcome associations.** We estimated associations using logistic regression fitted with generalized estimating equations (GEE), with an independence working correlation and robust standard errors clustered by agent to account for repeated observations. The safety analysis adjusted for underlying LLM, task, and command number, whereas the reliability analysis adjusted for underlying LLM and task.

**Burden and prevention.** We estimated potentially preventable fractions under specified hypothetical changes in behavioral exposures, conditional on causal assumptions, with confidence intervals from 1000 bootstrap resampling clustered by agent.

**Natural history.** We described how the frequency of behavioral exposures and adverse outcomes changed over agents' work lives, measured by task number. We also examined variation in outcomes between agents sharing the same underlying LLM.

**Subgroups and missing data.** No subgroup or interaction analyses were prespecified, and results by LLM are not reported as a ranking. No task records were missing in either cohort. No agent was lost to follow-up, and the protocols censor an agent at its last completed task if a technical fault stops it.

**Causal interpretation.** In both case studies we read an association as causal only under assumptions of adequate confounding control, correct temporal ordering, and valid measurement of exposure and outcomes. We adjusted for measured common causes of exposure and outcome, including the underlying LLM and task, and count behavioral exposure only when they precede the outcome. Nevertheless, unmeasured confounding cannot be ruled out. The data alone cannot test these assumptions. Thus, counterfactual estimates of potentially preventable events should be interpreted as conditional on these assumptions, rather than as demonstrated effects of an intervention.

**Reproducibility and reporting**

The complete study protocols, synthetic datasets and generators, task banks, sandbox and agent harness, scoring and analysis scripts, and full agent logs will be made available on publication. The studies are reported with reference to the STROBE guidelines for observational cohort studies, with the reporting checklist provided in the Supplementary Information.

The study is reported following the STROBE standard for observational studies (Table 1 and the checklist in the Supplementary Information). No human or patient data are used.

**Supplementary Table S1.** STROBE cohort reporting checklist for the epidemiologic studies of AI agents.

| STROBE item | Reporting requirement | Location in this study | Status |
|---|---|---|---|
| 1. Title and abstract | (a) Indicate the study design with a commonly used term in the title or abstract. (b) Provide | Main text, title and abstract; the title identifies a cohort design. | Addressed |
| | an informative and balanced summary of what was done and found. | | |
| 2. Background and rationale | Explain the scientific background and rationale for the investigation. | Main text, Introduction (capability-behavior distinction, epidemiologic analogy). | Addressed |
| 3. Objectives | State specific objectives, including any prespecified hypotheses. | Main text, Overview (three epidemiologic questions); Safety protocol, Objectives (primary and secondary aims). | Addressed |
| 4. Study design | Present key elements of study design early in the paper. | Main text, Overview and Methods; Safety protocol, Study design. | Addressed |
| 5. Setting | Describe the setting, locations and relevant dates, including periods of recruitment, exposure, follow-up and data collection. | Main text, Methods (virtual world, models); Safety protocol, Setting and Tasks. Actual enrollment and follow-up dates in Protocol. | Addressed |
| 6. Participants | (a) Give the eligibility criteria, the sources and methods of selection of participants, and the methods of follow-up. (b) For matched studies, give matching criteria and number of exposed and unexposed. | Main text, Methods (Models and deployment, Agent setup); Safety protocol, Participants and Study design. Part (b) is not applicable because there is no matching. | Addressed |
| 7. Variables | Clearly define all outcomes, | Main text, both case studies and | Addressed |
| | exposures, predictors, potential confounders and effect modifiers. Give diagnostic criteria, if applicable. | Methods; Safety protocol, Variables (operational definitions of curiosity and probing). | |
| 8. Data sources and measurement | For each variable of interest, give sources of data and details of methods of assessment. Describe comparability of assessment methods if there is more than one group. | Main text, Methods (Scoring); Safety protocol, Data sources and measurement (automatic logs, independent LLM rater, blinded human check); Supplement, validation of scoring. | Addressed |
| 9. Bias | Describe any efforts to address potential sources of bias. | Main text, Methods (at-risk periods, exposure classification, adjustment); Safety protocol, Bias and Mitigation. | Addressed |
| 10. Study size | Explain how the study size was arrived at. | Main text, sample sizes of both case studies; Safety protocol, Sample Size Estimation and Figure P1; Reliability protocol, sample size re-estimation to 106 agents. | Addressed |
| 11. Quantitative variables | Explain how quantitative variables were handled in the analyses. If applicable, describe which groupings were chosen and | Main text, Methods (exposure scales and analysis); Safety protocol, Variables and Statistical methods (curiosity status, command number, task | Addressed |
| | why. | number). | |
| 12. Statistical methods | (a) Describe all statistical methods, including those used to control for confounding. (b) Describe any methods used to examine subgroups and interactions. (c) Explain how missing data were addressed. (d) Explain how loss to follow-up was addressed. (e) Describe any sensitivity analyses. | Main text, Statistical analysis; Protocol, Statistical methods (GEE, adjustment, bootstrap) and Study design (censoring). Subitems (b) to (e) are stated in Subgroups, missing data and sensitivity. | Addressed |
| 13. Participants | (a) Report the numbers of individuals at each stage of the study. (b) Give reasons for non-participation at each stage. (c) Consider use of a flow diagram. | Main text, Results; safety cohort 200 agents and 3,600 tasks, reliability cohort 106 agents and 2,120 tasks, no censoring in either. Screening, exclusions included in Protocol. | Addressed |
| 14. Descriptive data | (a) Give characteristics of study participants and information on exposures and potential confounders. (b) Indicate the number of participants with missing data for each variable. (c) Summarize follow-up time. | Main text, Results (cohort size, LLM distribution, behavior frequencies); Safety protocol, Table P1. No missingness reported. | Addressed |
| 15. Outcome data | Report numbers of outcome events or | Main text, Results; 641 tasks with | Addressed |
| | summary measures over time. | probing, 631 tasks with a wrong or missing answer, natural-history summaries. | |
| 16. Main results | (a) Give unadjusted estimates and, if applicable, confounder-adjusted estimates and their precision. (b) Report category boundaries when continuous variables were categorized. (c) If relevant, consider translating relative risk into absolute risk for a meaningful time period. | Main text, Results (adjusted odds ratios, 95% confidence intervals, counterfactual fractions). | Addressed |
| 17. Other analyses | Report other analyses done, for example of subgroups and interactions, and sensitivity analyses. | Main text, Results (natural history, heterogeneity between agents of the same LLM, alternative adjustment for reliability). | Addressed |
| 18. Key results | Summarize key results with reference to study objectives. | Main text, Discussion. | Addressed |
| 19. Limitations | Discuss limitations of the study, taking into account sources of potential bias or imprecision. Discuss both direction and magnitude of any potential bias. | Main text, Discussion. | Addressed |
| 20. Interpretation | Give a cautious | Main text, | Addressed |
| | overall interpretation of results considering objectives, limitations, multiplicity of analyses, results from similar studies and other relevant evidence. | Discussion; the causal interpretation is to be moved into Methods. | |
| 21. Generalizability | Discuss the generalizability (external validity) of the study results. | Main text, Discussion (standardized virtual environments versus deployment). To be strengthened slightly. | Addressed |
| 22. Funding | Give the source of funding and the role of the funders for the present study and, if applicable, for the original study on which the present article is based. | Under Funding section. | Addressed |

## Code and data availability

Code and data will be made available on publication.

## Funding

This research is funded by Siteman Cancer Center and Washington University School of Medicine. The funder had no role in study design, data collection, analysis, interpretation, or manuscript preparation.

## Conflict of Interest

None reported related to this paper.

## References

1. Gao S, Fang A, Huang Y, Giunchiglia V, Noori A, Schwarz JR, et al. Empowering biomedical discovery with AI agents. Cell. 2024 Oct 31;187(22):6125-51.

2. Lin G, Jiang R, Elhadad N, Xu XO. A framework for longitudinal health AI agents. Nature health. 2026 May 22:1-0.
3. Chen L, Zhang Y, Feng J, Chai H, Zhang H, Fan B, et al. AI agent behavioral science. Humanities and Social Sciences Communications. 2026 Apr 28.
4. Mittal S. How Fast Do Agents Rot? An Empirical Study of Long-Horizon Degradation in LLM Agents for Production Decision-Making. arXiv preprint arXiv:2609.01660. 2026 Aug 31.
5. Al-Tawaha A, Gu S, Niu P, Jia R, Jin M. Remembering more, risking more: Longitudinal safety risks in memory-equipped LLM agents. arXiv preprint arXiv:2605.17830. 2026 May 18.
6. Wang Y, Goyal A, Chen Y, Sundaram H. State Contamination in Memory-Augmented LLM Agents. arXiv preprint arXiv:2605.16746. 2026 May 16.
7. Bogdanov I, Manakina O, Lung CH. Evaluation of Multi-Turn Consistency in LLM Agents: Survival Analysis and Failure-Rationale Taxonomy. arXiv preprint arXiv:2609.29508. 2026 Aug 26.
8. Rabanser S, Kapoor S, Kirgis P, Liu K, Utpala S, Narayanan A. Towards a science of AI agent reliability. arXiv preprint arXiv:2602.16666. 2026 Feb 18.
9. Liu Z. Memory Contagion: Cross-Temporal Propagation of Evaluator Bias via Agent Memory. arXiv preprint arXiv:2606.23195. 2026 Jun 22.
10. Bengio Y, Clare S, Prunkl C, Andriushchenko M, Bucknall B, Murray M, et al. International ai safety report 2026. arXiv preprint arXiv:2602.21012. 2026 Feb 24.
11. Embi PJ. Algorithmovigilance—advancing methods to analyze and monitor artificial intelligence–driven health care for effectiveness and equity. JAMA network open. 2021 Apr 15;4(4):e214622.
12. Dawber TR, Meadors GF, Moore Jr FE. Epidemiological approaches to heart disease: the Framingham Study. American Journal of Public Health and the Nations Health. 1951 Mar;41(3):279-86.
13. Colditz GA, Manson JE, Hankinson SE. The Nurses' Health Study: 20-year contribution to the understanding of health among women. Journal of women's health. 1997 Feb;6(1):49-62.
14. Colditz GA, Hankinson SE. The Nurses' Health Study: lifestyle and health among women. Nature Reviews Cancer. 2005 May 1;5(5):388-96.
15. Colditz GA, Winn DM. Criteria for the evaluation of large cohort studies: an application to the Nurses' Health Study. J Natl Cancer Inst. 2008;100(13):918-25.
16. Rothman KJ, Lash TL, Haneuse S, VanderWeele TJ. The scope of epidemiology. Modern epidemiology. 2021:3-15.
17. Colditz GA, Sellers TA, Trapido E. Epidemiology—identifying the causes and preventability of cancer?. Nature Reviews Cancer. 2006 Jan 1;6(1):75-83.
18. Parikh H, McCormick T, Johnson EK, Hickey L, Ranney ML, Mukherjee B. The epidemiology of artificial intelligence. Nature Health. 2026 Sep 15:1-0.
19. Von Elm E, Altman DG, Egger M, Pocock SJ, Gøtzsche PC, Vandenbroucke JP. The Strengthening the Reporting of Observational Studies in Epidemiology (STROBE) statement: guidelines for reporting observational studies. The lancet. 2007 Oct 20;370(9596):1453-7.
20. Ventures M. The state of generative AI in the enterprise. MenloVC. com. 2025 Dec.
21. OpenRouter. Model rankings by share of text requests, week of October 19, 2026. https://openrouter.ai/rankings (accessed 30 October 2026).
22. Colditz GA, Bonita R, Stampfer MJ, Willett WC, Rosner B, Speizer FE, et al. Cigarette smoking and risk of stroke in middle-aged women. New England Journal of Medicine. 1988 Apr 14;318(15):937-41.
23. Atlanta GA. The health consequences of smoking: 50 years of progress. A report of the surgeon general.[Google Scholar]. 2014 Jan.